\documentclass[aps,prl,reprint,superscriptaddress]{revtex4-2}
\usepackage{dcolumn}
\usepackage{graphicx}
\usepackage{amssymb}
\usepackage{amsmath}
\usepackage{annotate-equations}
\newcommand{\Rmnum}[1]{\uppercase\expandafter{\romannumeral #1}}

\begin{document}


\title{Emergence of a Flow-Assisted Casting Strategy for Olfactory Navigation via Memory-Augmented Reinforcement Learning}


\author{Changxu Zhao}
\affiliation{State Key Laboratory of Ocean Engineering, School of Ocean and Civil Engineering, Shanghai Jiao Tong University, Shanghai 200240, People’s Republic of China}

\author{Dongxiao Zhao}
\affiliation{State Key Laboratory of Ocean Engineering, School of Ocean and Civil Engineering, Shanghai Jiao Tong University, Shanghai 200240, People’s Republic of China}

\author{Xin Bian}
\affiliation{State Key Laboratory of Fluid Power and Mechatronic Systems, Department of Engineering Mechanics, Zhejiang University, Hangzhou 310027, People’s Republic of China}

\author{Gaojin Li}
\email{gaojinli@sjtu.edu.cn}
\affiliation{State Key Laboratory of Ocean Engineering, School of Ocean and Civil Engineering, Shanghai Jiao Tong University, Shanghai 200240, People’s Republic of China}



\date{\today}

\begin{abstract}
In dynamic flow fields, various animals exhibit remarkable odor search capabilities despite relying on stochastic detections. Interestingly, there exists an optimal time window for integrating these detections that maximizes search efficiency. To understand the underlying mechanism, we investigate the navigation performance of Reinforcement Learning (RL) agents in unsteady flows under varying memory lengths and flow conditions. Without any predefined models, the agents develop a flow-assisted casting strategy and adaptively adjust both the geometry of their search trajectories and the concentration threshold for initiating casting to maximize the success rate. The agent's average speed toward the odor source exhibits a non-monotonic dependence on memory length, which can be explained by the ``sector-search'' model.
\end{abstract}

\maketitle

\section{Introduction}

Understanding odor searching in animals within dynamic environments is important for deciphering their survival-critical behaviors \cite{vickers2000mechanisms, murlis1992odor, reddy2022olfactory} and for advancing robotic navigation in low-detectability environments across diverse applications, such as targeted therapy \cite{liebchen2019optimal,wang2024tracking}, pollutant monitoring \cite{dunbabin2012robots, de2022innovative}, and deep-sea exploration \cite{li2023bioinspired}. In this challenging task, agents must detect fluctuating, intermittent odors against a low-concentration background and resume the tracking of sparse, tortuous odor filaments after a temporary loss of signal \cite{celani2014odor,carde2008navigational,riffell2008physical}. They often also exploit ambient flow currents to facilitate navigation toward the target \cite{gunnarson2021learning,jiao2025sensing,stupski2024wind}. However, the underlying physics and neural mechanisms for olfactory navigation remain not fully understood.

Searching for a target of unknown location can be modeled as a partially observable Markov decision process (POMDP) \cite{spaan2012partially}, which requires solving a nonlinear Bellman equation, yet finding its exact optimal solution is computationally intractable \cite{kurniawati2022partially,kochenderfer2022algorithms}. Depending on how the observation history shapes the planning or learning policies, various approximate solvers have been developed, including rule-driven approaches \cite{balkovsky2002olfactory}, finite state controllers \cite{verano2023olfactory}, infotaxis and its extensions \cite{vergassola2007infotaxis,barbieri2011trajectories,masson2013olfactory}, sampling-based methods \cite{heinonen2023optimal,heinonen2025optimal,heinonen2025exploring}, and reinforcement learning \cite{loisy2022searching,reddy2022sector,loisy2023deep,singh2023emergent,rando2025q}. The length of the history, or the memory for the agent, plays a crucial role in determining search efficiency. Chemotaxis and other single-step policies with zero memory can only handle simple scenarios. Model-based methods, such as infotaxis and sampling-based approaches, internally encode a perfect memory to update the belief about possible source locations, often leading to near-optimal trajectories \cite{vergassola2007infotaxis, loisy2022searching, heinonen2025optimal, heinonen2025exploring}. For RL agents, finite memories are sufficient for odor search \cite{reddy2022sector}, and an optimal length exists that maximizes  both the success rate and average speed of reaching the target \cite{masson2013olfactory, grunbaum2015spatial, rando2025q}. It was recently shown that the optimal memory length matches the typical duration of no odor detection \cite{rando2025q} and later confirmed by measurements of odor-evoked neuron activities in \textit{Drosophila} \cite[]{kathman2025neural}. However, previous studies typically neglect the kinetics of agent in dynamic environments, and some rely on predefined models, such as the free-energy based model \cite{masson2013olfactory} or the sector-search model \cite{reddy2022sector}, to guide odor search. It remains unclear how the performance of a search kinetics of a model-free RL agent with finite memory compares to these theoretical models.





In this work, we introduce a finite memory length to an RL agent using a Bootstrapped Random Update (BRU) algorithm and quantify its influence on odor search performance across a large parameter space. By analyzing the average behavior of search trajectories, we identify the key features of the agent's navigation strategy in a dynamic flow field. The agents' average speed is further analyzed through a geometric analysis of the ``sector-search'' casting model to explain the existence of an optimal memory length.

\begin{figure}
\includegraphics[width=\linewidth]{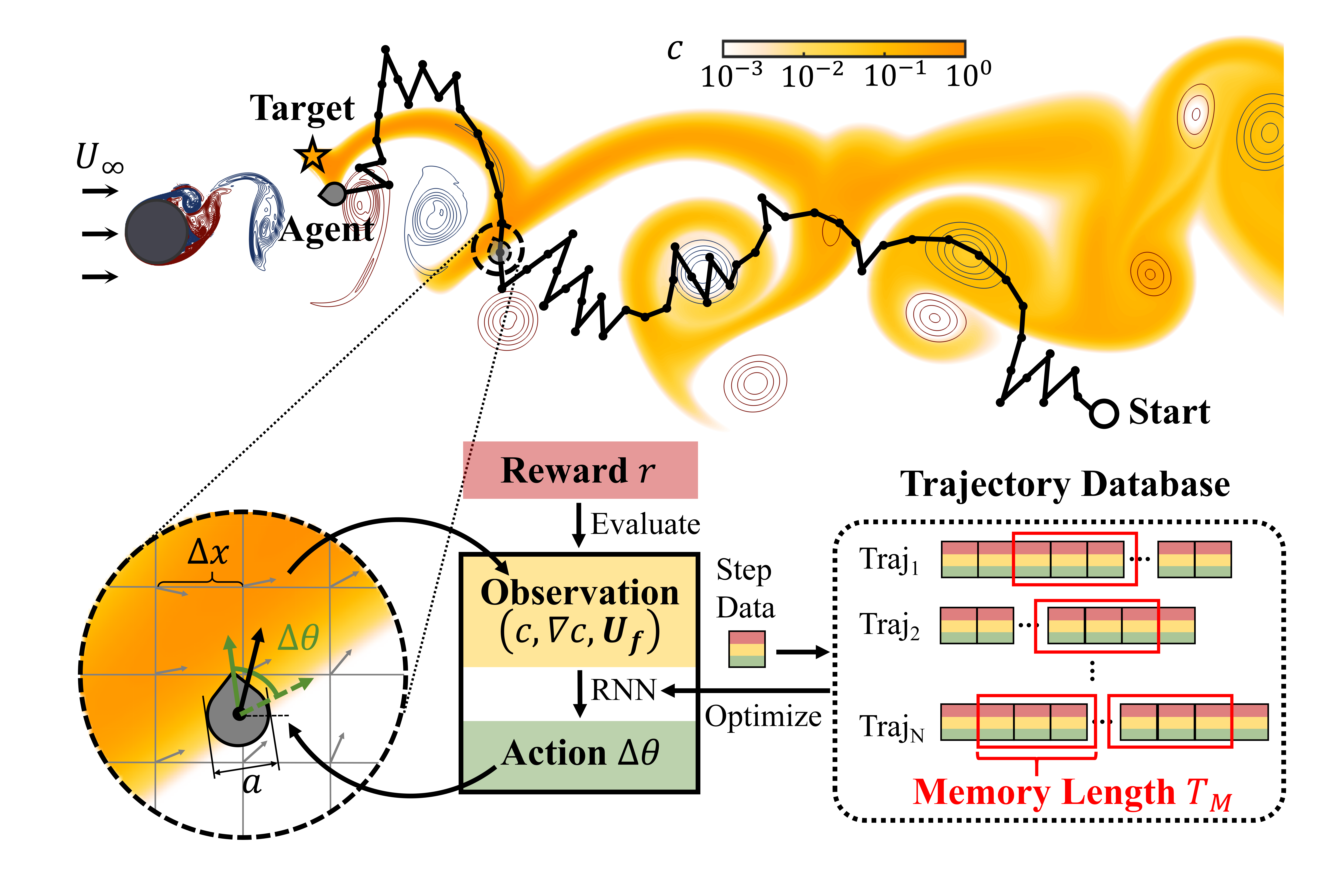}
\caption{Schematic of the odor source searching of a finite-memory RL agent trained by the BRU strategy in an unsteady flow field. At each time step, the agent observes the local concentration and fluid velocity $(c, \nabla c, \boldsymbol{U}_f)$, adjusts its heading angle $\theta$ via a RNN policy, and updates its position based on self-propulsion combined with the background flow. Stepwise trajectory data, including observations, actions, and rewards, are stored in a history database, from which memory segments with fixed length $T_M$ are randomly sampled to train the action policy.}
\label{fig:model_algorithm}
\end{figure}



As shown in Fig.\ref{fig:model_algorithm}, we consider a finite-memory RL agent tracking odor plumes in an unsteady flow field with concentration $c(\boldsymbol{x}, t)$ and fluid velocity $\boldsymbol{U}_f(\boldsymbol{x}, t)$. The field data are obtained from direct simulations of von Kármán vortex streets behind a cylinder, with Reynolds number $Re=U_\infty D/\nu=100\sim5000$ and Schmidt number $Sc=\nu/\kappa=10^{-2}\sim1$, where $U_\infty$ is the inflow velocity, $D$ is the cylinder diameter, and $\nu$ and $\kappa$ are the kinematic viscosity and odor diffusivity of the fluid, respectively. At each time step $t$, the agent detects local concentration and fluid velocity $(c, \nabla c, \boldsymbol{U}_f)$ as inputs, sampled at frequency $f$ from a discretized field with grid resolution $\Delta x=0.1D$, and then updates its moving direction by $\Delta\theta$ as the action. The agent's speed $U_a$ is assumed constant, and its position is updated based on both its own velocity and the local fluid velocity, $\boldsymbol{X}_a(t+\Delta t)=\boldsymbol{X}_a(t)+\Delta t[\boldsymbol{U}_a+\boldsymbol{U}_f(\boldsymbol{X}_a, t)]$, where $\Delta t=1/f$ is the time step. 


We use a Random-updated Recurrent Soft Actor-Critic (RRSAC) algorithm \cite{mywork2026} to embed sufficient complexity into the agent model while adopting a relatively simple flow configuration and including local fluid velocity in the detection information to reduce training difficulty. Recent studies on navigation \cite{jiao2025sensing} and odor search \cite{mywork2026} have shown that the velocity information does not qualitatively affect the agent's performance. At each step, the agent’s reorientation angle $\Delta\theta$ is sampled from a Gaussian distribution whose mean and variance are determined by a Recurrent Neural Network (RNN) policy. Through the BRU strategy, the RNN is continuously updated using memory segments of finite length $T_M$, each comprising the agent's observations, actions, and rewards collected along its trajectory. In each learning episode, the location of the odor source and the agent's starting position and time are randomly assigned to prevent the agent from memorizing positional information. See Supporting Information for details on the RL framework and training settings \cite{supplementary}.

\begin{figure*}
\includegraphics[width=\linewidth]{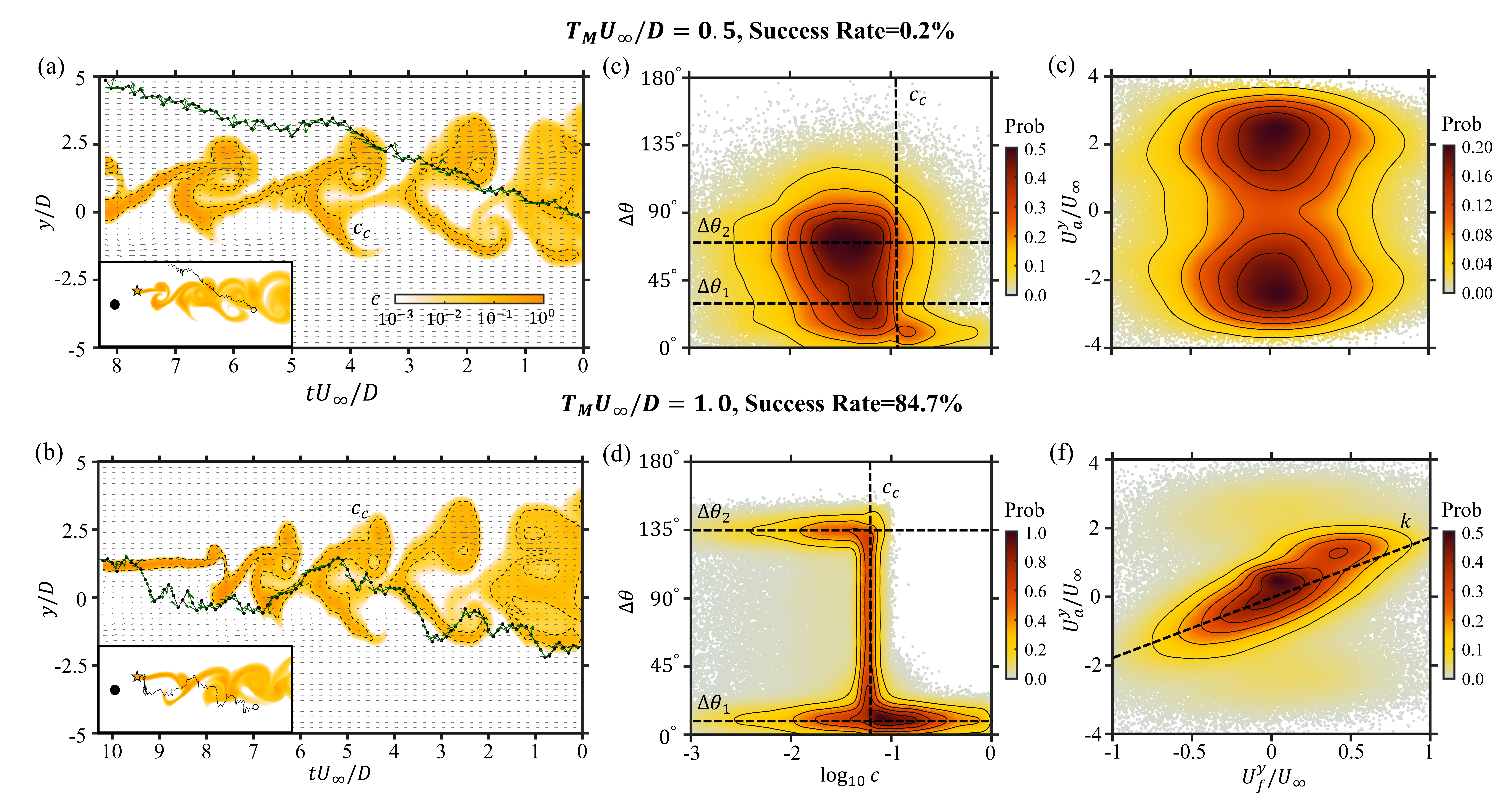}
\caption{Emergence of flow-assisted casting strategy with increased memory length $T_M$. (a,b) Spatiotemporal evolution of agent's vertical position $Y_a$ (black line) and heading direction (green arrows) for $T_MU_\infty/D=0.5$ and $1.0$, along with the background concentration field (color) and flow velocity (gray arrows). Insets show the agents' trajectories. (c-f) Joint probability distributions in the spaces of (c,d) turning angle $\Delta\theta$ versus local concentration $\log_{10}c$, and (e,f) agent's vertical velocity $U_a^y$ versus flow vertical velocity $U_f^y$. The remaining parameters are $Re=5000, Sc=0.1, U_a=3U_{\infty}$, and $f=10U_{\infty}/D$.}
\label{fig:hybrid strategy}
\end{figure*}


Each agent typically reaches a plateau of accumulated reward after approximately 2000 training episodes. Thereafter, 1000 independent search trials are conducted for analysis. Fig. \ref{fig:hybrid strategy}(a,b) show the typical spatiotemporal evolution of the agent's $y$-position and the local flow field for memory length $T_{M}U_\infty/D=0.5$ and $1$ (note the time increases from right to left). The remaining parameters are $Re=5000, Sc=0.1, U_a=3U_{\infty}$, and $f=10U_\infty/D$, and the success rates of the two agents in reaching the source are 0.2\% and 84.7\%, respectively. Consistent with observations of many insects \cite{murlis1992odor,ryohei1992self,mafra1994fine,carde2008navigational} and terrestrial animals \cite{vickers2000mechanisms,reddy2022olfactory,khan2012rats}, the agents exhibit ``zigzaggin'' behavior regardless of their actual searching performance (see trajectories in insets). The short-memory agent exhibits high-frequency, small-amplitude oscillations. In contrast, the long-memory agent develops a robust hybrid trajectory composed of zigzagging paths with increasing amplitude in low-concentration regions and relatively straight paths in high-concentration regions. Meanwhile, it strategically adjusts its orientation to follow the vertical component of the background flow to enhance the locomotion efficiency, demonstrating a flow-assisted casting strategy.




This strategy becomes more evident through statistical analysis. In Fig.\ref{fig:hybrid strategy}(c-f), we plot the joint probability distributions of the agent's state in the $\Delta\theta-\log_{10}c$ and $U_a^y-U_f^y$ spaces across all search trials, where $\Delta\theta=\cos^{-1}\left(
\frac{\boldsymbol{U}_a(t)\cdot\boldsymbol{U}_a(t+\Delta t)}{|\boldsymbol{U}_a(t)||\boldsymbol{U}_a(t+\Delta t)|}\right)$ is the turning angle of the agent between two consecutive time steps, and $U_a^y, U_f^y$ are the $y$-components of the agent's self-propulsion velocity and the local flow velocity, respectively. Interestingly, the two agents develop different strategies in responding to concentration and flow velocity. Below a concentration threshold $c_c$, the turning angle of the long-memory agent exhibits a clear bimodal distribution, favoring either a small adjustment or a large reorientation. Near the threshold, the turning angles become more uniformly distributed, and for $c>c_c$, the agent predominantly persists in its heading direction. For the agent with $T_MU_\infty/D=1$, the joint PDF is strongly concentrated at $c_c\simeq0.064$ and two turning angles $\Delta\theta_1\simeq9^\circ$ and $\Delta\theta_2\simeq135^\circ$. For the short-memory agent with $T_MU_\infty/D=0.5$, the PDF is spread much more broadly, with $c_c\simeq0.107, \Delta\theta_1\simeq30^\circ$, and $\Delta\theta_2\simeq72^\circ$. In Fig.\ref{fig:hybrid strategy}(e, f), both plots show probability spread out over the range of $U_f^y/U_{\infty}\in[-1, 1]$, suggesting that the agents tend to stay inside the vortical wake region to enhance movement, as also observed in point-to-point navigation \cite{gunnarson2021learning}. However, the patterns differ markedly depending on the memory length. The short-memory agent tends to have $U_a^y/U_{\infty}\simeq\pm2$ irrespective of $U_f^y$, whereas for the long-memory agent, $U_a^y$ shows an approximately linear correlation with the local crossflow velocity $U_f^y$, with slope $k=U_a^y/U_f^y\simeq1.86$.

\begin{figure}
\includegraphics[width=\linewidth]{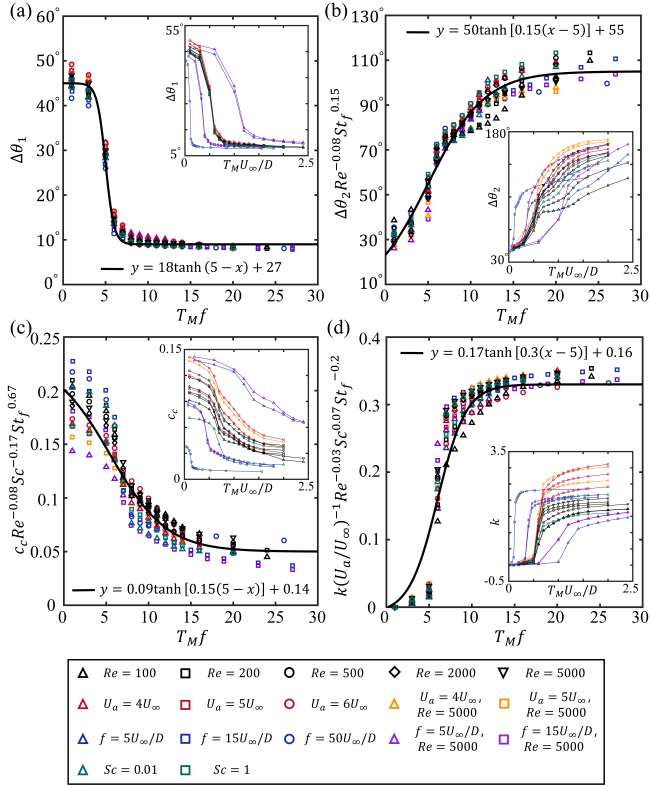}
\caption{Universal dependence of (a) small adjusting angle $\Delta\theta_1$, (b) large casting angle $\Delta\theta_2$, (c) concentration threshold $c_c$, and (d) velocity ratio $k$ on the memory length $T_M$ for agents under different configurations and flow conditions. The default parameters are $Re=500, Sc=0.1, U_a=3U_\infty$, and $f=10U_\infty/D$.}
\label{fig:characteristic}
\end{figure}

The above results remain robust across diverse flow conditions ($Re, Sc$) and agent configurations ($U_a/U_{\infty}, f$, and $T_M$) \cite{supplementary}. In each case, the agent exhibits a ``zigzagging'' trajectory characterized by the parameters $\Delta\theta_1,\Delta\theta_2,c_c$ and $k$, and flow-assisted casting strategy becomes more pronounced with increasing memory length. In Fig. \ref{fig:characteristic}(a), the small adjusting angle $\Delta\theta_1$ collapses onto a single curve when plotted against the memory length scaled by the detection frequency, $T_Mf$, instead of the time scale of the flow field, $D/U_\infty$ (see the inset). It decreases drastically from $\sim45^\circ$ to $\sim10^\circ$ above a critical memory $T_{M,c}f\simeq6$, after which the decrease becomes much slower. In Fig.\ref{fig:characteristic}(b)-(d), the results are rescaled as power functions of $Re, Sc, U_a/U_\infty$, and the detection Strouhal number $St_f=fD/U_a$ to best collapse them onto single curves by minimizing the sum of squared residuals. Terms with absolute power coefficient smaller than 0.02 are neglected. As the memory length increases, the casting angle $\Delta\theta_2$ rapidly rises from approximately $45^\circ$ to $100^\circ-170^\circ$, meaning that the agents follow casting trajectories with pronounced sharp turns. The critical concentration $c_c$ decreases monotonically with $T_M$, suggesting that agents become more confident in moving along near-straight trajectories with a long memory. The velocity ratio $U_a^y/U_f^y$ increases monotonically with $T_M$, implying more efficient exploitation of the background fluid velocity by longer-memory agents.

\begin{figure}
\includegraphics[width=\linewidth]{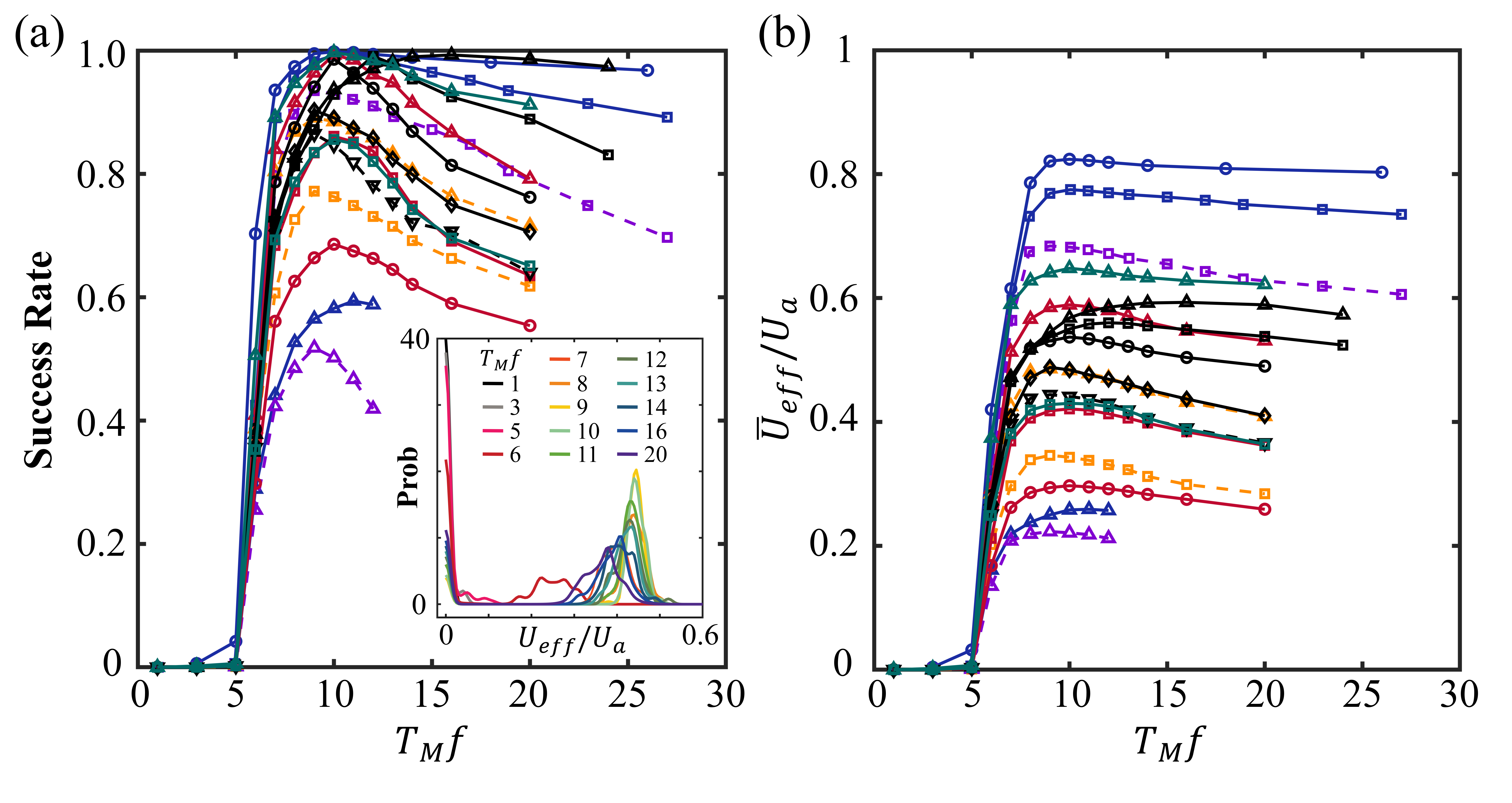}
\caption{Non-monotonic dependence of search performance on memory length $T_M$. (a) Success rate and (b) average effective speed $\bar{U}_{eff}$ versus $T_M$ for agents under different conditions (same legend as in Fig. \ref{fig:characteristic}). The inset in (a) shows the probability distributions of $U_{eff}/U_a$ for different $T_M$, with $Re=5000, Sc=0.1, U_a=3U_{\infty}$, and $f=10U_\infty/D$.}
\label{fig:ueff}
\end{figure}

The agent's ability to locate the odor source depends critically on its memory length. As shown in Fig.\ref{fig:ueff}(a), the success rate exhibits a non-monotonic dependence on memory length. Below the critical memory length $T_{M,c}\simeq6/f$, the success rate remains close to zero. Once exceeds this threshold, the agent's performance rises steeply, reaching a maximum at an optimal value $T_M^*\sim10/f$. Depending on the parameters, the maximum success rate ranges from approximately $50-60\%$ for agents with low detection frequency in high-$Re$ flows to about $99\%$ for cases with high frequency and low $Re$. Above the optimal memory length, the success rate slowly declines. This trend is further illustrated by the probability distribution of the effective speed $U_{eff}$, defined as the initial agent–source distance divided by the arrival time, for agents at $Re = 5000$ (see the inset, and see \cite{supplementary} for the distributions under different conditions). Agents with short and long memory lengths exhibit distinctly different distributions of $U_{eff}$. For small $T_M$, most trials fail, and the distribution is sharply peaked at $U_{eff}=0$. As $T_M$ increases, this peak quickly diminishes while a second peak, representing successful trials, emerges at higher speeds and progressively shifts rightward. Beyond $T_M^*\simeq10/f$, this peak shifts back to the left. Near the optimal memory length, the effective speed exhibits the narrowest distribution, reflecting robust odor search by the agent. In Fig. \ref{fig:ueff}(b), the average effective speed $\bar{U}_{eff}$ across all search trials shows the same trends as the success rate. For relatively difficult search trials (large $Re, Sc$ and low $f$), the average speed is about $20-30\%$ of the largest possible speed if neglect the background flow. For simpler cases (small $Re, Sc$ and high $f$), the speed reaches approximately $70-80\%$ of the straight-line speed, demonstrating the effectiveness of the search strategy.








\begin{figure}
\includegraphics[width=0.9\linewidth]{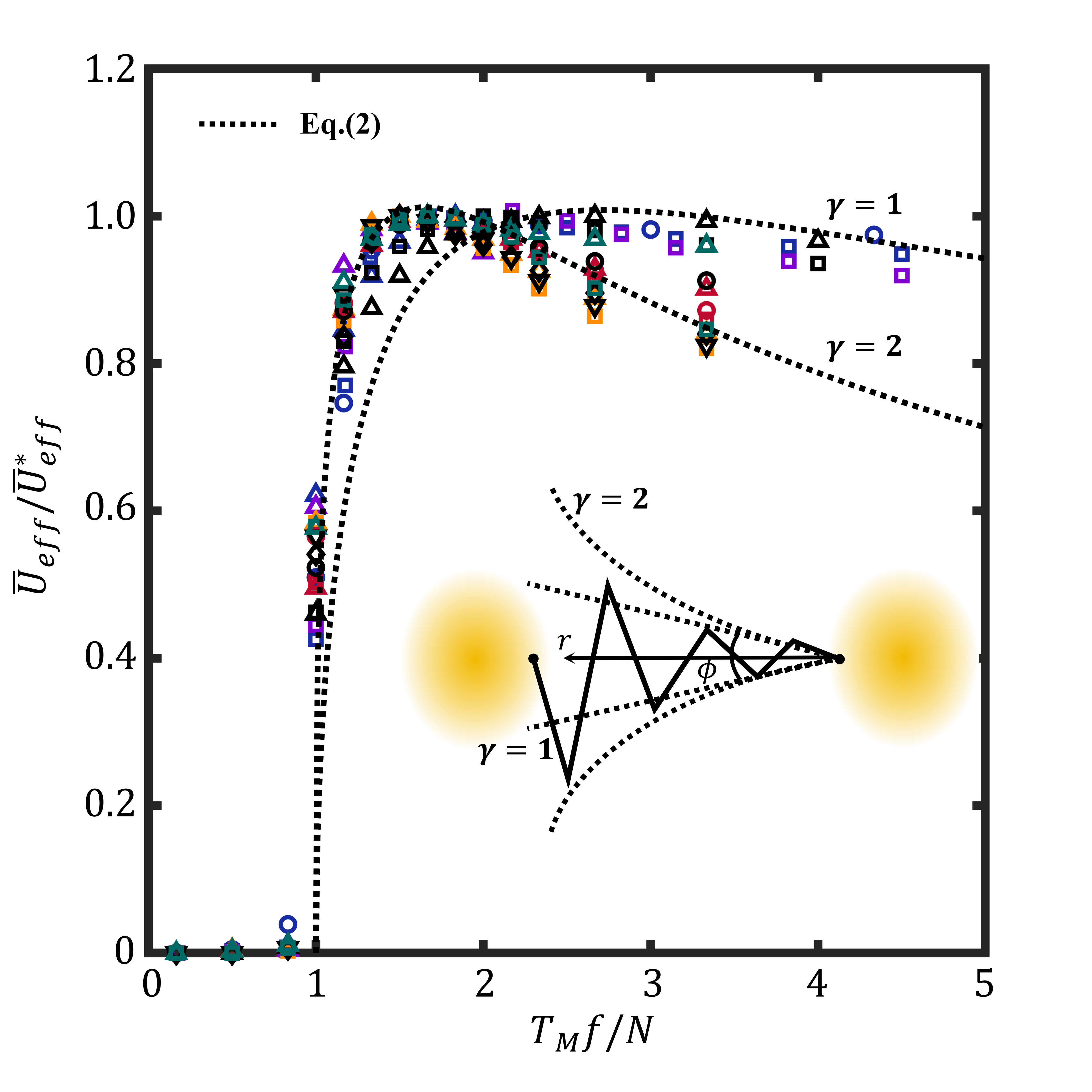}
\caption{Comparison of our simulated normalized average speed with the ``sector-search'' model. The inset illustrates the casting behavior between two 
consecutive concentration detections.}
\label{fig:rescaled}
\end{figure}

To further understand the influence of memory length on search speed, we transfer the ``sector-search'' model, originally proposed for static odor trail tracking \cite{reddy2022sector}, to the context of odor search in dynamic environments. As shown in the inset of Fig.~\ref{fig:rescaled}, the agent initiates an expanding cast trajectory once losing odor detection. The angle of the scanning area grows as $d\phi/dt=fa/r$, where $f$ is the detection frequency, $a$ is the agent size, and $r=\bar{U}_{eff}t$ is the radial distance from the location where the track is lost. The opening angle of the scanning area is estimated as $\phi=af\ln(Tf/N)/\bar{U}_{eff}$, with $T$ the time duration between two consecutive odor detections, and $N$ the number of casting segments. On the other hand, the angle $\phi$ reflects the uncertainty of the agent's heading direction and is assumed to grow with the radial distance as $\phi=(\bar{U}_{eff}T/l)^\gamma$, where the constants $l$ and $\gamma$ determine the growth rate of the angle. Equating these two equations, and to prevent the agent from completely forgetting the odor track, the agent's memory length should match the typical non-detection duration, $T_M=T$, the average effective speed is then solved as 
\begin{equation}\label{ec:11}
\bar{U}_{eff}=\left(\frac{al^{\gamma}f}{T_M^{\gamma}}\ln(\frac{T_Mf}{N})\right)^{\frac{1}{1+\gamma}}.
\end{equation} 
The speed $\bar{U}_{eff}=0$ for $T_Mf\leq N$, and it reaches its maximum $\bar{U}_{eff}^*=((al^\gamma f^{1+\gamma})/(e\gamma N^\gamma))^{1/(1+\gamma)}$ at $T_M^*f=e^{1/\gamma}N$. From Fig.~\ref{fig:ueff}, we estimate $N=6$, meaning that the agent completes three full casting cycles on average in low-concentration regions, consistent with the observations in Fig.~\ref{fig:hybrid strategy}(b). We use it as the only fitting parameter and rescale all data by the corresponding maximum speed. Fig.~\ref{fig:rescaled} compares the simulated results of the normalized speed $\bar{U}_{eff}/\bar{U}_{eff}^*$ with the theoretical prediction,
\begin{equation}\label{ec:12}
\frac{\bar{U}_{eff}}{\bar{U}_{eff}^*}=\left(e\gamma\ln(\frac{T_Mf}{N})/\left(\frac{T_Mf}{N}\right)^\gamma\right)^{\frac{1}{1+\gamma}}.
\end{equation} 
For relatively challenging search trials, our results agree well with the prediction for $\gamma=2$. From this result, we can calculate the optimal memory length as $T_M^*f\simeq9.89$, which agree well with our results shown in Fig.~\ref{fig:hybrid strategy}(b). For easier trials, $\bar{U}_{eff}$ decreases more slowly ($\gamma\simeq1$) at larger $T_M$, indicating self-learning of a more conservative casting strategy under less challenging conditions. Below the optimal memory length $T_M^*$, the agents outperform the theoretical prediction, indicating the benefit of exploring the background flow field. 

The existence of optimal memory lengths agrees with our finding in Fig.~\ref{fig:ueff}, which shows that that agents can only reach the odor source when their memory is sufficiently long. Furthermore, increasing $T_M$ decreases $c_c$ and increases $\Delta\theta_2$, indicating a shift toward a more conservative strategy that postpones  casting and promotes broader search upon losing detection, at the expense of forward progression. From the perspective of RL-strategy training, a shorter memory allows for more diverse training data and random sampling from each trajectory, improving algorithm efficiency and model stability, while it truncates the temporal context, hindering the agent's ability to learn long-range environmental correlations.

In summary, we studied the dynamics of odor search for a finite-memory-augmented RL agent in a highly unsteady flow field. Relying solely on locally sensed concentration and flow velocity information, the agent develops a hybrid strategy that combines the well-known cast-and-surge behavior \cite{david1983finding,mafra1994fine} with a tendency to move along the cross-flow direction, which is commonly observed for navigation in unsteady flows \cite{krishna2022finite,biferale2019zermelo}. Depending on the complexity of the search task, the agent adaptively adjusts its scanning angle and the concentration threshold for casting to optimally balance exploration and exploitation. Furthermore, we showed that the non-monotonic dependence of the agent’s speed on memory length can be well described by the ``sector-search'' model.

\nocite{narvekar2016source,narvekar2020curriculum,hausknecht2015deep,shankar2016reinforcement,karkus2017qmdp,ng1999policy,eck2016potential}

\begin{acknowledgments}
\textbf{Acknowledgment} This work is supported by the Natural Science Foundation of China (grant nos 12372264, 12102258, 12202270) and the Natural Science Foundation of Shanghai (grant no. 23ZR1430800).
\end{acknowledgments}

\nocite{*} 
\bibliography{references}

\end{document}


\title{Supplementary Information: \\ Emergence of a Flow-Assisted Casting Strategy for Olfactory Navigation via Memory-Augmented Reinforcement Learning}


\author[1]{Changxu Zhao}
\author[1]{Dongxiao Zhao}
\author[2]{Xin Bian}
\author[1]{Gaojin Li \thanks{gaojinli@sjtu.edu.cn}}
\affil[1]{State Key Laboratory of Ocean Engineering, School of Ocean and Civil Engineering, Shanghai Jiao Tong University, Shanghai 200240, People’s Republic of China}
\affil[2]{State Key Laboratory of Fluid Power and Mechatronic Systems, Department of Engineering Mechanics, Zhejiang University, Hangzhou 310027, People’s Republic of China}

\maketitle


\section{Supplementary details on model and algorithm}

\subsection{Training Environment and Randomization Settings}

\begin{figure}[htbp]
\centering
\includegraphics[width=0.9\linewidth]{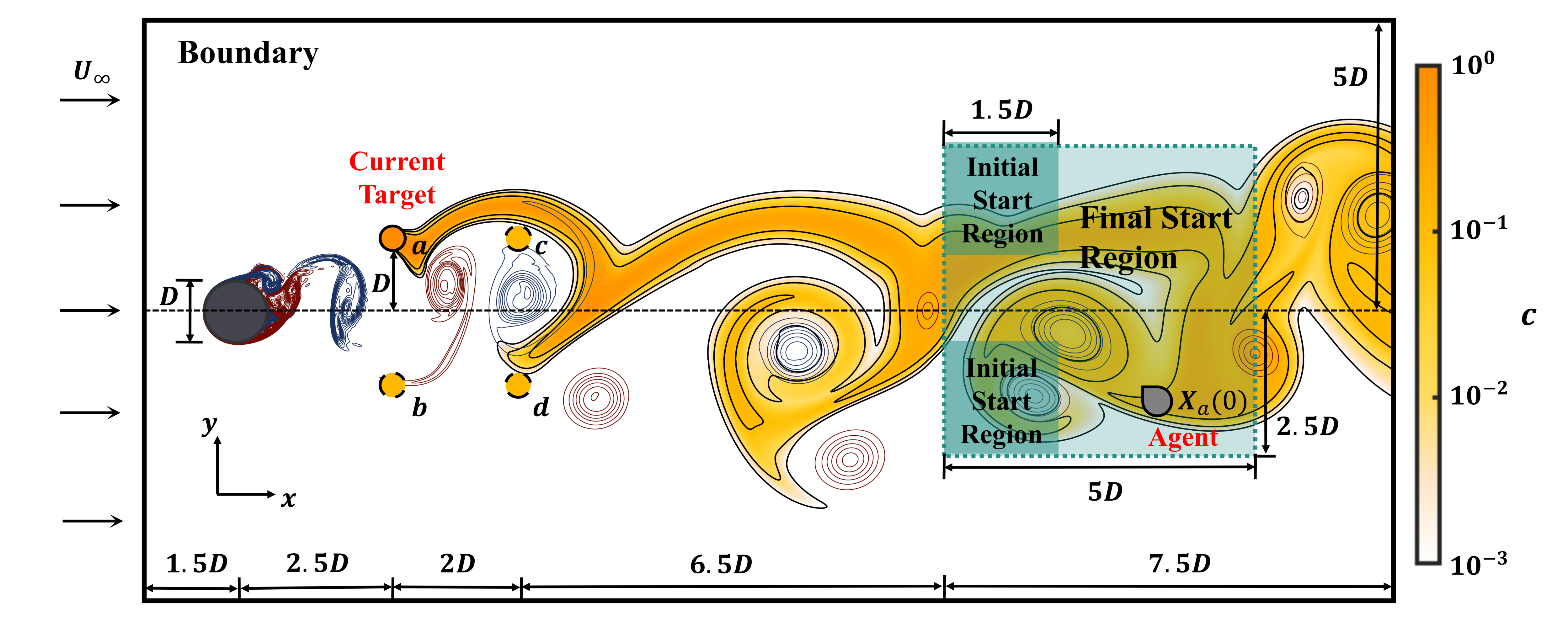}
\caption{Training environment for an agent searching a odor source in an unsteady flow. The background shows the concentration field, with contours shown at $10^{-3},10^{-2},10^{-1}$. A target region of radius $0.2D$ is randomly placed at one of four locations, labeled $a$ through $d$. The light and dark cyan areas indicate the agent's start regions for early and later training episodes, respectively. Each swimmer begins an episode with randomized initial position, orientation, and start time. A success episode is defined as the agent entering the target region, i.e., $|\boldsymbol{X}_{a}-\boldsymbol{X}_{target}|<\delta$.}
\label{fig:simulation environment}
\end{figure}

\quad Here, we provide an overview of our model and algorithm for odor search. Complete algorithmic details can be found in our recent work \cite{mywork2026}. To provide the agent with immediate observation inputs $(c, \nabla c, \boldsymbol{U}_f)$, we numerically simulate the concentration field in a von Kármán vortex street by solving the Navier-Stokes equation and advection-diffusion equation. As shown in Figure \ref{fig:simulation environment}, the computational domain is $\Omega_f=[0,20D]\times[0,10D]$, with a cylinder at $(1.5D,5D)$, where $D$ is the diameter. The chemical source is a circular disk of radius $\delta=0.2D$ randomly located at one of the four fixed positions $\boldsymbol{X}_{target}$ (labeled $a$-$d$). The source emits chemical substances at a constant rate $J$, contributing a source term $J\pi\delta^{2} h_{\delta}(\boldsymbol{x}-\boldsymbol{X}_{target})$ to the concentration $c(\boldsymbol{x},t)$, where $h_{\delta}(\boldsymbol{x} - \boldsymbol{X}_{target}) = \mathbf{1}_{\{ \|\boldsymbol{x} - \boldsymbol{X}_{target}\| \leq \delta \}}$ is a circular unit step function. The full simulation runs for $160D/U_{\infty}$, where $U_\infty$ is the velocity of the coming flow. Only the final $60D/U_{\infty}$ is used to train the agent, as the flow becomes stable by this stage. This duration substantially exceeds typical search time $\sim10D/U_{\infty}$ and typical non-detection intervals, ensuring sufficient exploration. For each perception step, the background field data is mapped onto a uniform $200\times 100$ grid with spacing $\Delta x = 0.1D$ using natural neighbor interpolation based on Delaunay triangulation. The agent starts randomly from an initial point $\boldsymbol{X}_{a}(0)$ within the start region at an initial time $t_{0}U_{\infty}/D\in[0,5]$. At each step, the agent updates its position and obtains observations by interpolating from nearby grid points.

\subsection{Curriculum Task Planning}

\quad To facilitate the acquisition of robust source-searching strategies in complex flows, we use a curriculum learning approach \cite{narvekar2016source,narvekar2020curriculum} that progressively adjusts the difficulty of training tasks. During the early training phase, the start regions are confined to relatively small regions (dark cyan areas in Figure \ref{fig:simulation environment}). For every $n_{e}$ episodes, the rectangular start area enlarges proportionally to include more distant and complex flow zones (light cyan regions in Figure \ref{fig:simulation environment}). The curriculum level is defined as $l=\left \lfloor n/n_{e} \right \rfloor$, where $\left \lfloor \cdot \right \rfloor$ denotes the floor function, and $n$ is the current learning episode. This level dynamically sets the size of the start region: $\text{Width} = 1.5D + 3.5Dl/\left \lfloor N/n_{e} \right \rfloor$ and $\text{Height} = 1.5D + Dl/\left \lfloor N/n_{e} \right \rfloor$, where the total number of episodes $N$ is set to 10,000 in our research.

\subsection{RNN-based Partial Observation Markov Decision Process}

\quad Odor searching is a typical Partial Observation Markov Decision Process (POMDP), as the agent cannot directly perceive the true state (e.g. its relative position to the target). Formally, a POMDP could be characterized by a 7-tuple $(\mathcal{S,A,T,R,\gamma,O},\Omega)$, where the variables are as follows:

\begin{itemize}
    \item $\mathcal{S}$ is the state space, which is the set containing all possible states of the system (agent and environment). Ideally, a state $s \in \mathcal{S}$ contains all information needed for optimal sequential decision-making.
    \item $\mathcal{A}$ is the action space, which is the set of all possible actions the agent can perform.
    \item $\mathcal{T}:\mathcal{S} \times \mathcal{A} \times \mathcal{S} \rightarrow [0,1]$ is the state transition probability, where $T(s'|s,a)$ denotes the probability of transitioning to the next state $s'$ after taking action $a$ in state $s$.
    \item $\mathcal{R}: \mathcal{S} \times \mathcal{A} \rightarrow \mathbb{R}$ is the reward function, where $r = \mathcal{R}(s,a)$ represents the immediate scalar reward the agent receives after taking action $a$ in state $s$.
    \item $\gamma \in [0,1]$ is the discount factor, which balances immediate versus future rewards. A smaller $\gamma$ prioritizes short-term returns, while a larger $\gamma$ encourages long-term planning.
    \item $\mathcal{O}$ is the observation space, which is the set of all possible observations the agent can perceive.
    \item $\Omega\mathcal{:S \times O}\rightarrow [0,1]$ is the emission probability, where $\Omega(o \mid s)$ represents the probability that the agent perceives observation $o$ in state $s$.
\end{itemize}

At each timestep $t$, the agent interacts with the environment from an unknown state $s_{t}$. Under partial observability, it maintains a belief state $b \in \mathcal{B}$ as an estimate of the true state, where $\mathcal{B}$ denotes the belief space. The agent selects an action $a_{t}$ according to its policy $\pi:\mathcal{B} \times \mathcal{A}\rightarrow [0,1]$, receives a scalar reward $r_{t} = R(s_{t},a_{t})$, and transitions to a new state governed by a conditional probability $T(s_{t+\Delta t}|s_t,a_t)$. Notably, both the reward function $R$ and the state transition function $T$ are intrinsic to the environment and unknown to the agent. The agent's goal is therefore to learn a policy $\pi$ that maximizes the expected discounted reward sum $\mathbb{E}\left[ \sum_{t} \gamma^{t} r_{t} \right]$, through trial-and-error interactions with the environment.

In model-free RL, a recurrent neural network (RNN) is introduced to model the belief state $b_t$ by encoding historical observations $\{o_0,\cdots,o_{t-\Delta t}\}$ into a hidden state $h_{t}\in\mathcal{H}$ \cite{hausknecht2015deep,shankar2016reinforcement,karkus2017qmdp}. The pair $(h_t,o_t)$ captures the agent's full history and current uncertainty, allowing it to infer spatiotemporal dynamics and learn a source searching policy $\pi(\cdot \mid o_{t},h_{t})$ that selects a heading angle adjustment $\Delta \theta_{t}$ as action at each step.

\subsection{Random-updated Recurrent Soft Actor-Critic (RRSAC) Algorithm}

\subsubsection{Bootstrapped Random Update}

\begin{figure}[htbp]
\centering
\includegraphics[width=0.9\linewidth]{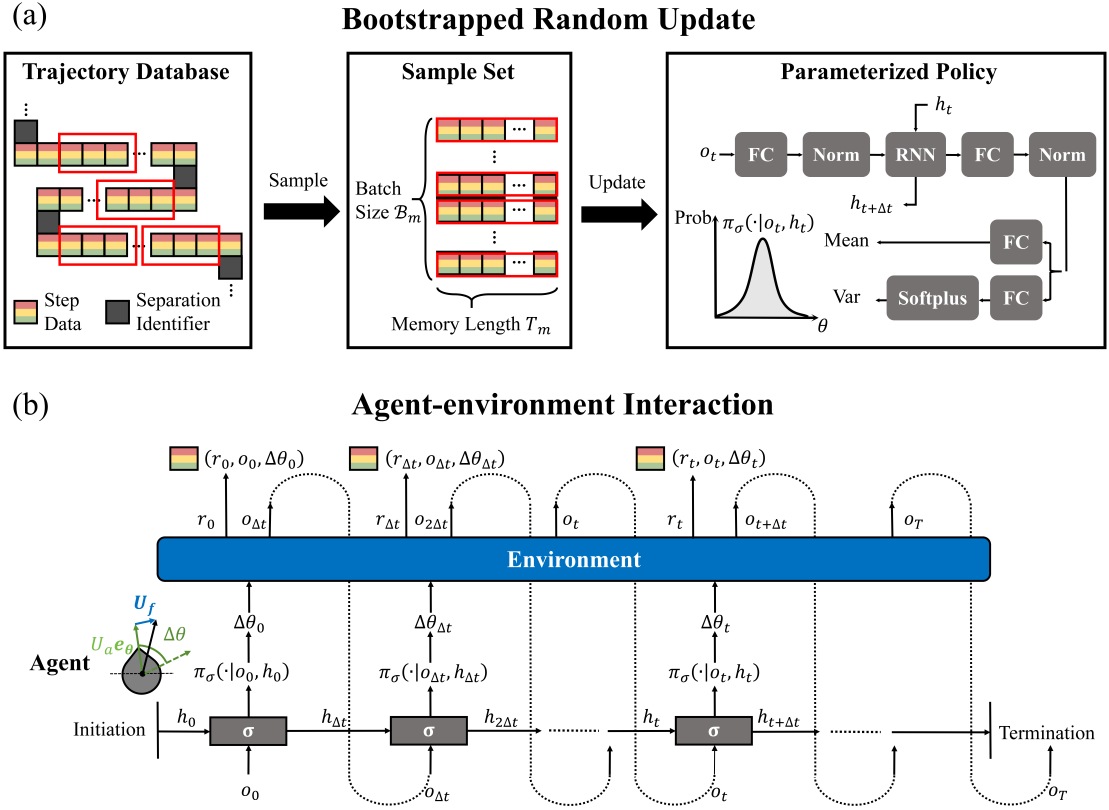}
\caption{Overview of the RRSAC algorithm. (a) BRU technique: Step data is stored sequentially in a trajectory database. Each update samples a batch of fixed-length $T_M$ trajectory segments to train the parameterized policy $\pi_{\sigma} (\cdot \mid o_{t}, h_{t})$. FC represents a fully connected layer, Norm denotes layer normalization, and Softplus is an activation function. (b) Agent–environment interaction: Starting from an initial condition, the hidden state $h_{t}$ is iteratively updated during exploration. At each step, the policy uses hidden state $h_{t}$ and observation $o_{t}$ to produce an action distribution, from which a heading adjustment $\Delta \theta_t$ is sampled. The environment then returns a reward $r_{t}$ and a new observation $o_{t+\Delta t}$. Reward, observation, and action are stored as step data to form the trajectory database.}
\label{fig:Neural network architecture}
\end{figure}

\quad To model the agent’s finite memory, a Bootstrapped Random Update (BRU) technique is employed to update the odor search policy. This policy is parameterized by a neural network $\pi_{\sigma}( \cdot)$ comprising an RNN unit. Stepwise trajectory data are stored sequentially in the trajectory database, with episode identifiers to separate different searching trails. During the update phase, a batch $\mathcal{B}_{m}$ of exploration episodes is sampled with replacement. For each sampled episode, a random starting point is chosen, and policy gradients are computed over a fixed-length window $T_{M}$, representing the agent’s memory length. As illustrated in Figure \ref{fig:Neural network architecture}(a), this BRU scheme ensures that the training samples are approximately independent and identically distributed (i.i.d.), and allows efficient use of partial trajectories.

\subsubsection{RRSAC Scheme}

\quad The agent employs an RRSAC scheme for odor search, integrating recurrent mechanisms into the policy of the Soft Actor-Critic (SAC) algorithm. This policy is trained with the BRU technique using trajectory data from agent-environment interactions, as shown in Figure \ref{fig:Neural network architecture}(b). At each time step $t$, the policy receives the previous hidden state $h_{t}$ and the current observation $o_{t}$. An internal RNN unit $f_{R}$ updates the hidden state via $h_{t+\Delta t}=f_{R}(o_{t},h_{t})$, and the policy outputs a stochastic distribution $\pi_{\sigma} (\cdot \mid o_{t},h_{t})$, typically Gaussian. An action $a_{t}$ is then sampled from this distribution and executed in the environment, yielding a reward $r_{t}$ and the next observation $o_{t+\Delta t}$ for the subsequent step. These interactions sequentially form a POMDP trajectory $\tau =(h_{0},o_{0},a_{0},...,h_{t},o_{t},a_{t},...)$, where $\rho_{\pi_{\sigma}}$ denotes its distribution induced by $\pi_{\sigma}$. To encourage exploration, RRSAC optimizes the policy using a maximum entropy objective $\sigma^{*}=\argmax_{\sigma}\mathbb{E}_{\tau \sim \rho_{\pi_{\sigma}}} \left[ \sum_{t} \gamma^{t}\left(r(o_{t},h_{t},a_{t}) +\xi \mathcal{E}(\pi _{\sigma}(\cdot \mid o_{t},h_{t}))\right)\right]$. Here $\mathcal{E}(\pi_{\sigma}(\cdot \mid o_{t},h_{t}))$ is policy entropy and $\xi$ is a temperature coefficient. We employ Adam stochastic optimization and burn-in technique for updating network parameters.

\subsubsection{Reward Setting}

\quad Our reward function $r_t$ combines an original component with a potential-based shaping term. The original reward consists of a chemotaxis bonus $\alpha(\boldsymbol{X}_{a}(t+\Delta t)-\boldsymbol{X}_a(t)) \cdot \nabla c/|\boldsymbol{X}_{a}(t+\Delta t)-\boldsymbol{X}_a(t)||\nabla c|$, a time penalty $-\beta\Delta t$, and corrections for boundary crossing, successful arrival, and timeouts. To accelerate learning, we incorporate a reward shaping term derived from a potential function $\Phi(s_{t})=-\left \| \boldsymbol{X}_a(t)-\boldsymbol{X}_{target} \right \|/U_a$, formulated as $F(s_{t},s_{t+\Delta t})=\gamma\Phi(s_{t+\Delta t})-\Phi(s_{t})$. This approach improves learning efficiency without altering the original optimal policy, as documented in the literature \cite{ng1999policy,eck2016potential}. An annealing mechanism, $\mu e^{-\zeta n/N}F(s_{t},s_{t+\Delta t})$, is also applied to gradually diminish the shaping effect. The coefficients $\{\alpha,\beta,\mu,\zeta\}$ balance these terms to optimize exploration and policy learning.

\subsection{Evaluation Environment Setting}

\quad We evaluate each trained agent over 1000 independent trials in its training environment with initial randomization. Notably, the initial point is chosen from the final start region of larger size shown in Figure \ref{fig:simulation environment}. During evaluation, the parameterized policy is fixed. Each trial uses trained parameters for hidden state updates and stochastic action generation. The parameters setting are shown in tables \ref{tab:simulation_settings} and \ref{tab:algorithm_training}.

\clearpage

\subsection{Parameters Setting}

\begin{table}[H]
    \centering
    \caption{Parameters for simulation and reward setting}
    \label{tab:simulation_settings}
    \begin{tabular}{ll}
        \toprule
        Parameter description & Values \\
        \midrule
        Flow simulation time step       & 0.01$D/U_{\infty}$ \\
        Chemical diffusivity $\kappa$            & 0.02$DU_{\infty}$ \\
        Kinematic viscosity $\nu$       & 0.002$DU_{\infty}$ \\
        Concentration emitting rate $J$ & 0.1$U_{\infty}/D$ \\
        Maximum exploration time        & 55$D/U_{\infty}$ \\
        $\{\alpha,\beta,\mu,\zeta\}$    & $\{4,1,800,2.5\}$ \\
        Successful arrival reward                 & 3000 \\
        Crossing boundary penalty       & -200 \\
        Timeout penalty                 & -2000 \\
        \bottomrule
    \end{tabular}
\end{table}

\begin{table}[H]
    \centering
    \caption{Parameters for algorithm and training}
    \label{tab:algorithm_training}
    \begin{tabular}{ll}
        \toprule
        Parameter description & Values \\
        \midrule
        Model layer initializations (RNN, FC) & (Normal, Orthogonal) \\
        Hidden layers                         & 32,32,128,32,32 \\
        Activation function                   & Leaky ReLU \\
        Norm                                  & Layer Norm \\
        RNN                                   & LSTM \\
        Learning rate of actor & 1e-4 \\
        Learning rate of Q-function  & 1e-3 \\
        Learning rate of temperature $\xi$    & 1e-4 \\
        Total number of episodes $N$          & 1e4 \\
        Environment update step length $n_e$       & 50$\Delta t$ \\
        Training update step length        & 10$\Delta t$ \\
        Burn-in length            & 10$\Delta t$ \\
        Discount factor $\gamma$              & 0.99 \\
        Soft update parameter           & 0.002 \\
        Buffer capacity                       & 50000 \\
        Update beginning threshold            & 500 \\
        Batch size $\mathcal{B}_{m}$          & 64 \\
        \bottomrule
    \end{tabular}
\end{table}

\clearpage

\section{Extended Results}

This section presents additional results, including the joint PDF in $\Delta \theta-\log_{10} c$ and $U_{a}^{y}-U_{f}^{y}$ spaces, spatiotemporal evolution diagrams, and PDF of $U_{eff}/U_a$ under different conditions. Two animations illustrate representative search trajectories for agents at $Re=500$ and $Re=5000$, respectively. The remaining parameters are $Sc=0.1,U_a=3U_{\infty}$, and $f=10U_{\infty}/D$.




\subsection{Joint PDF in $\Delta \theta-\log_{10} c$ and $U_{a}^{y}-U_{f}^{y}$ Spaces for Different $T_M$}

\begin{figure}[htbp]
\centering
\includegraphics[width=0.95\linewidth]{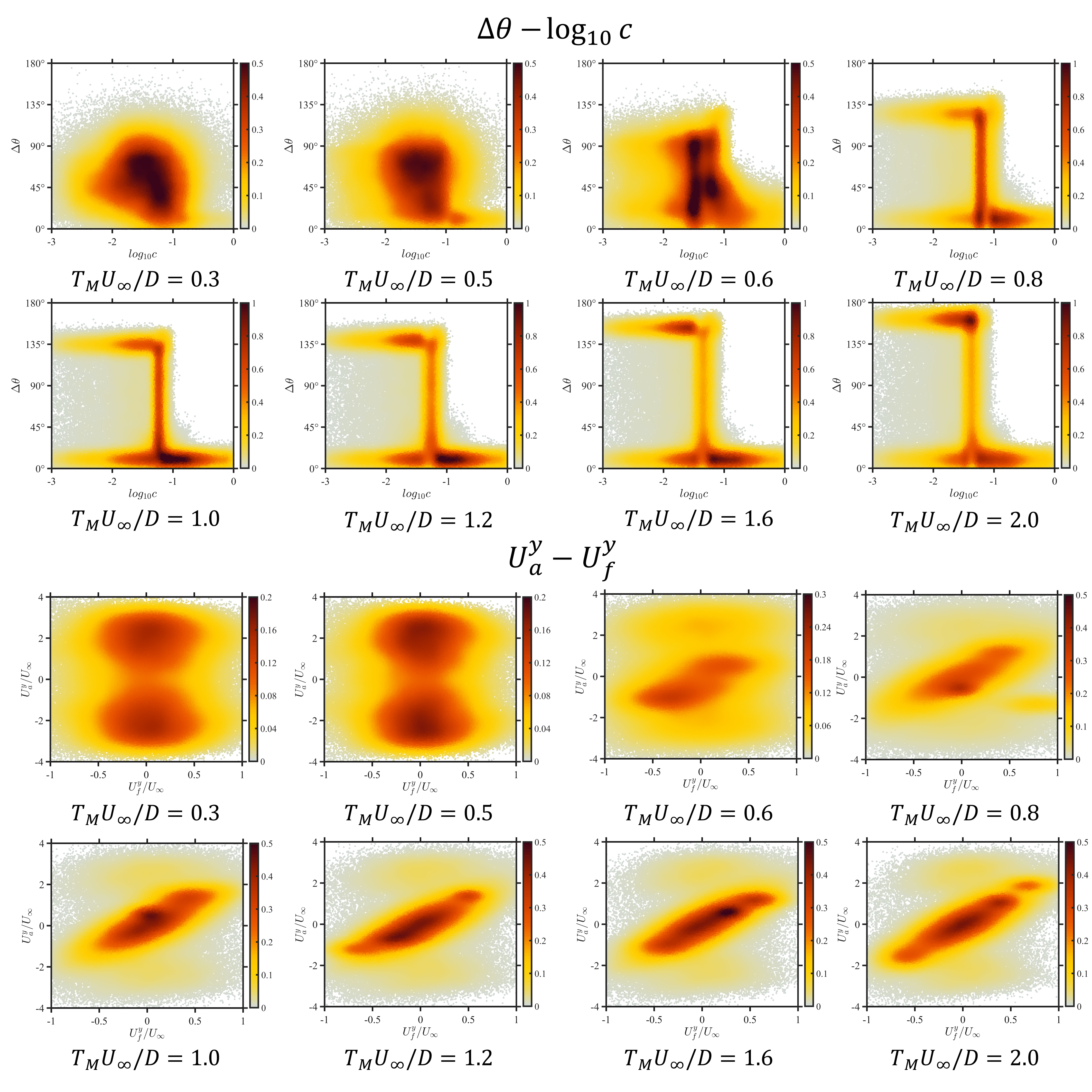}
\caption{Joint probability density distribution of the agent’s state in $\Delta \theta-\log_{10} c$ and $U_{a}^{y}-U_{f}^{y}$ space for memory lengths $T_M$ ranging from $0.3D/U_{\infty}$ to $2D/U_{\infty}$. The remaining parameters are $Re=5000,Sc=0.1,U_a=3U_{\infty},f=10U_{\infty}/D$.}
\label{fig:supplement_s4}
\end{figure}

\quad To better illustrate how the odor search strategy evolves with memory length $T_M$, figure \ref{fig:supplement_s4} presents the joint probability density distributions of the agent’s state in $\Delta \theta-\log_{10} c$ and $U_{a}^{y}-U_{f}^{y}$ space for $T_M$ ranging from $0.3D/U_{\infty}$ to $2D/U_{\infty}$. As explained in the main text, in the $\Delta \theta-\log_{10} c$ space the distribution of turning angle $\Delta \theta$ relative to local concentration undergoes a "bimodal–uniform–unimodal" pattern separated by the concentration threshold $c_c$. In the $U_{a}^{y}$–$U_{f}^{y}$ space, the distribution reveals an approximately linear relationship between the $y$-component of the agent’s self-propulsion velocity $U_a^y$ and the local flow velocity $U_f^y$. As the memory length $T_M$ increases, the distribution becomes more concentrated and displays a pronounced flow-assisted casting strategy.

\subsection{Spatiotemporal Evolution Diagrams under Different Conditions}

\begin{figure}[htbp]
\centering
\includegraphics[width=1\linewidth]{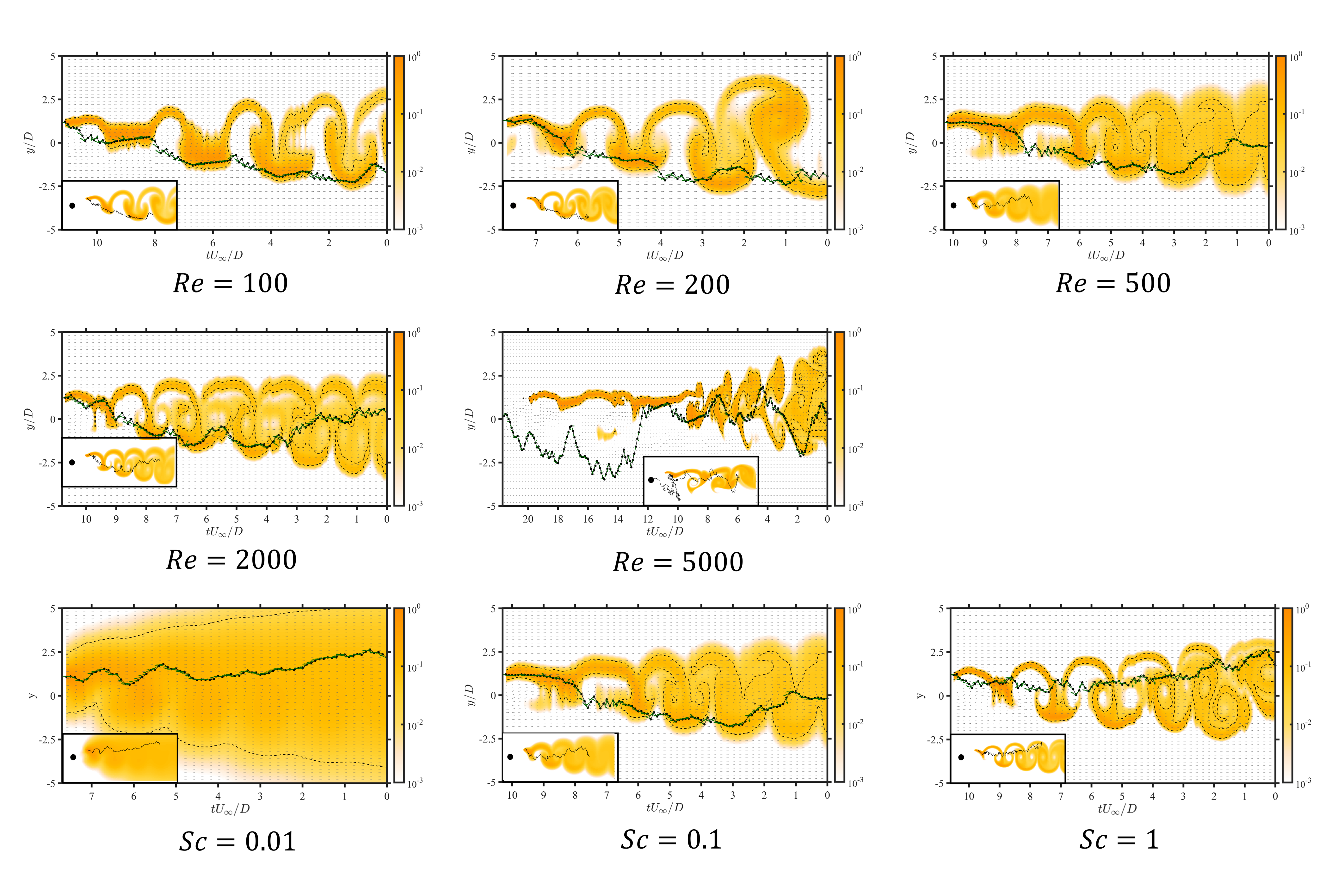}
\caption{Spatiotemporal evolution of the agent’s vertical position $Y_a$ under different flow conditions $Re$ and $Sc$. Every agent is equipped with the optimal memory length $T_M^*$, and the default parameters are $Re=500,Sc=0.1,U_a=3U_{\infty},f=10U_{\infty}/D$.}
\label{fig:supplement_s5}
\end{figure}

\begin{figure}[htbp]
\centering
\includegraphics[width=1\linewidth]{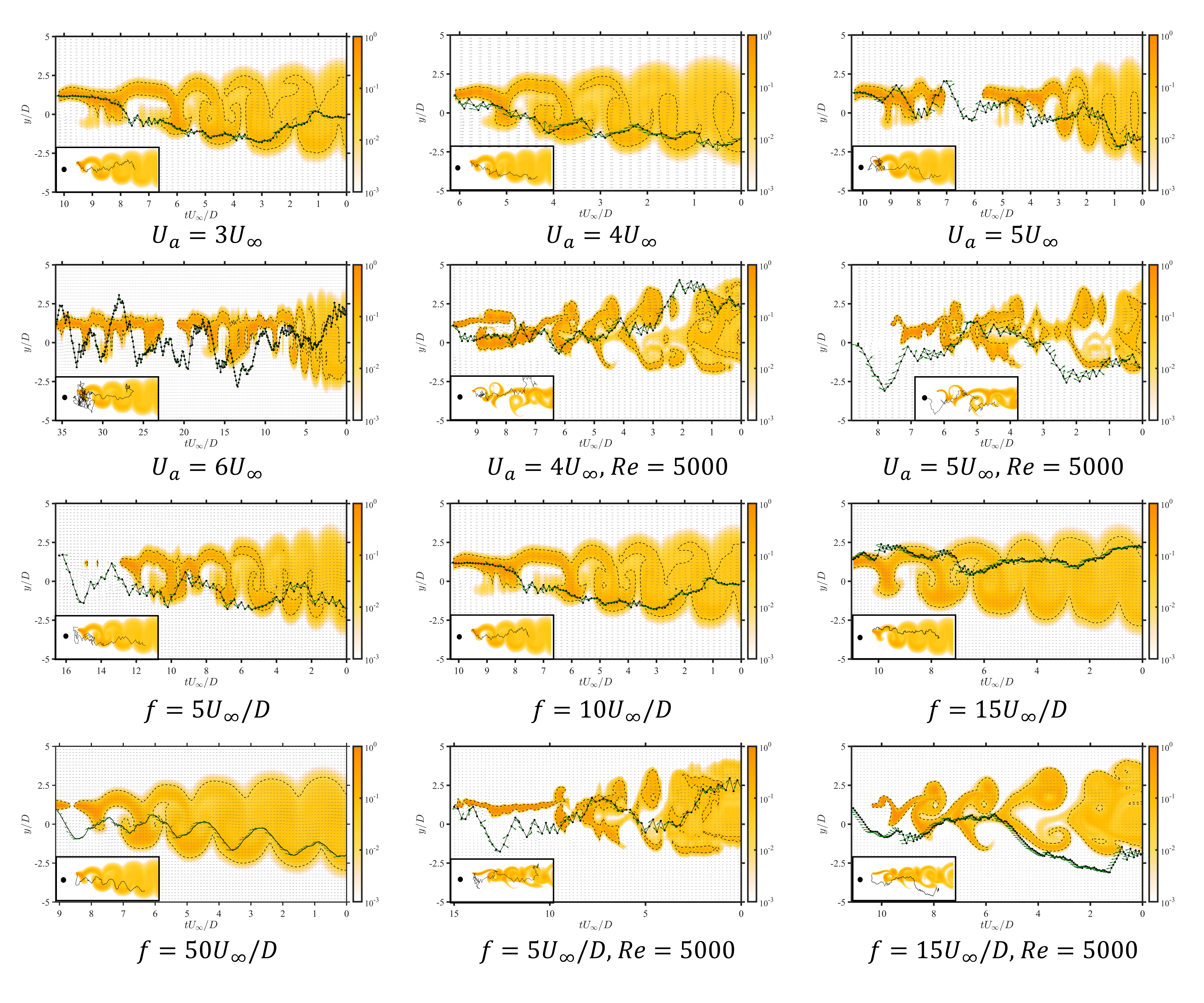}
\caption{Spatiotemporal evolution of the agent’s vertical position $Y_a$ under different agent configurations $U_a/U_{\infty}$ and $f$. Every agent is equipped with the optimal memory length $T_M^*$, and the default parameters are $Re=500,Sc=0.1,U_a=3U_{\infty},f=10U_{\infty}/D$.}
\label{fig:supplement_s6}
\end{figure}

\quad The flow-assisted casting strategy remains robust across different conditions and adapts well to the complexity of the odor search task. Figures \ref{fig:supplement_s5} and \ref{fig:supplement_s6} show the spatiotemporal evolution of the agent’s vertical position $Y_a$ under various flow conditions ($Re,Sc$) and agent configurations ($U_a/U_{\infty}, f$), with the concentration threshold $c_c$ marked by a black dotted contour. Below $c_c$, the agent follows zigzagging paths of increasing amplitude, whereas above $c_c$, it follows relatively straight paths. In challenging search scenarios (characterized by large $Re, Sc, U_a/U_\infty$ and low $f$), the agent adopts a large concentration threshold $c_c$ and zigzags frequently toward high-concentration regions, resulting in tortuous trajectories (see insets) and low search speed. In simpler cases (small $Re, Sc, U_a/U_\infty$ and high $f$), the agent adopts a a small $c_c$ and yields smoother trajectories and shorter search times.

\subsection{Joint PDF in $\Delta \theta-\log_{10} c$ and $U_{a}^{y}-U_{f}^{y}$ Spaces under Different Conditions}

\begin{figure}[htbp]
\centering
\includegraphics[width=0.95\linewidth]{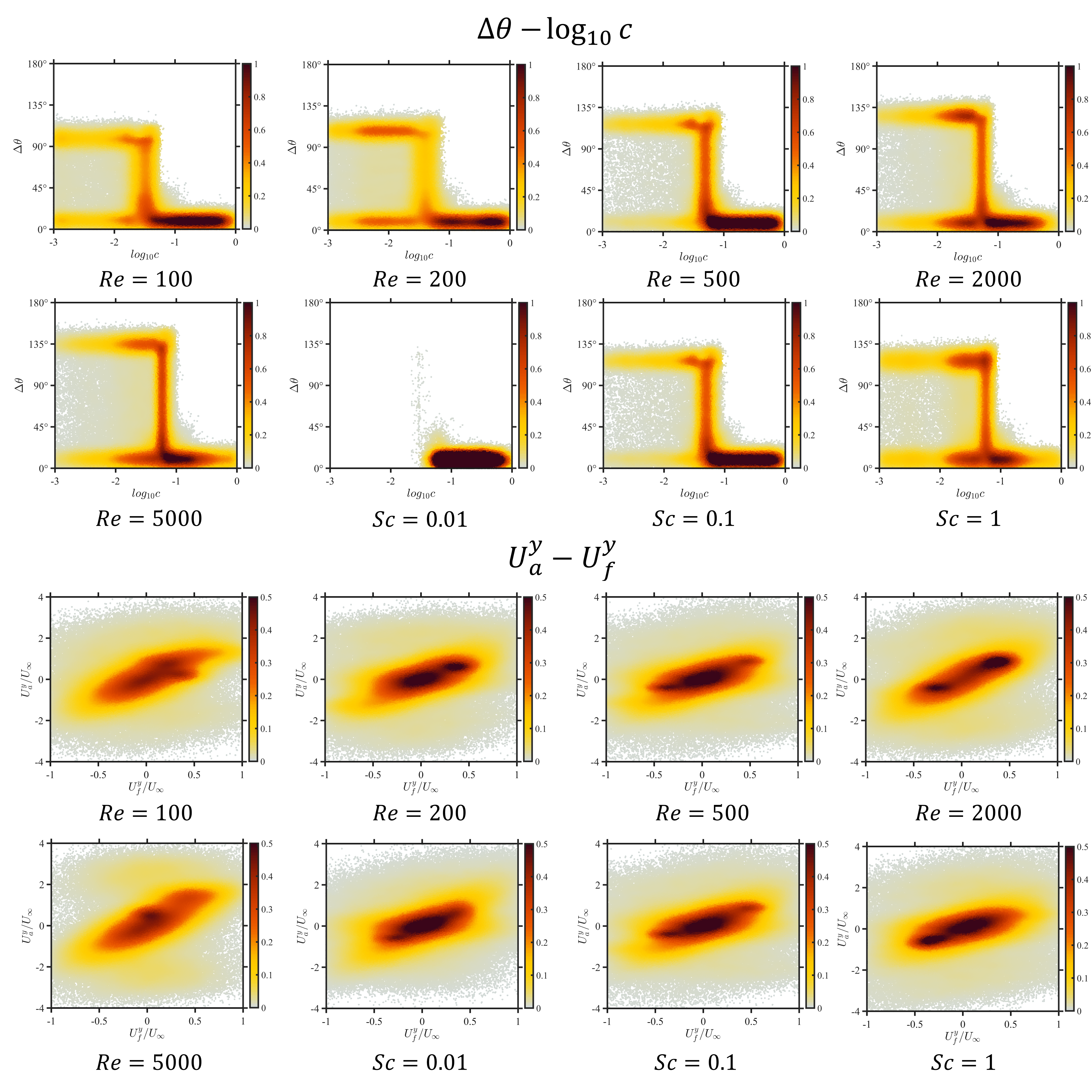}
\caption{Joint probability density distribution of the agent’s state in $\Delta \theta-\log_{10} c$ and $U_{a}^{y}-U_{f}^{y}$ space with optimal memory length $T_M^*$ for different $Re$ and $Sc$. The default parameters are $Re=500,Sc=0.1,U_a=3U_{\infty},f=10U_{\infty}/D$.}
\label{fig:supplement_s7}
\end{figure}

\begin{figure}[htbp]
\centering
\includegraphics[width=0.95\linewidth]{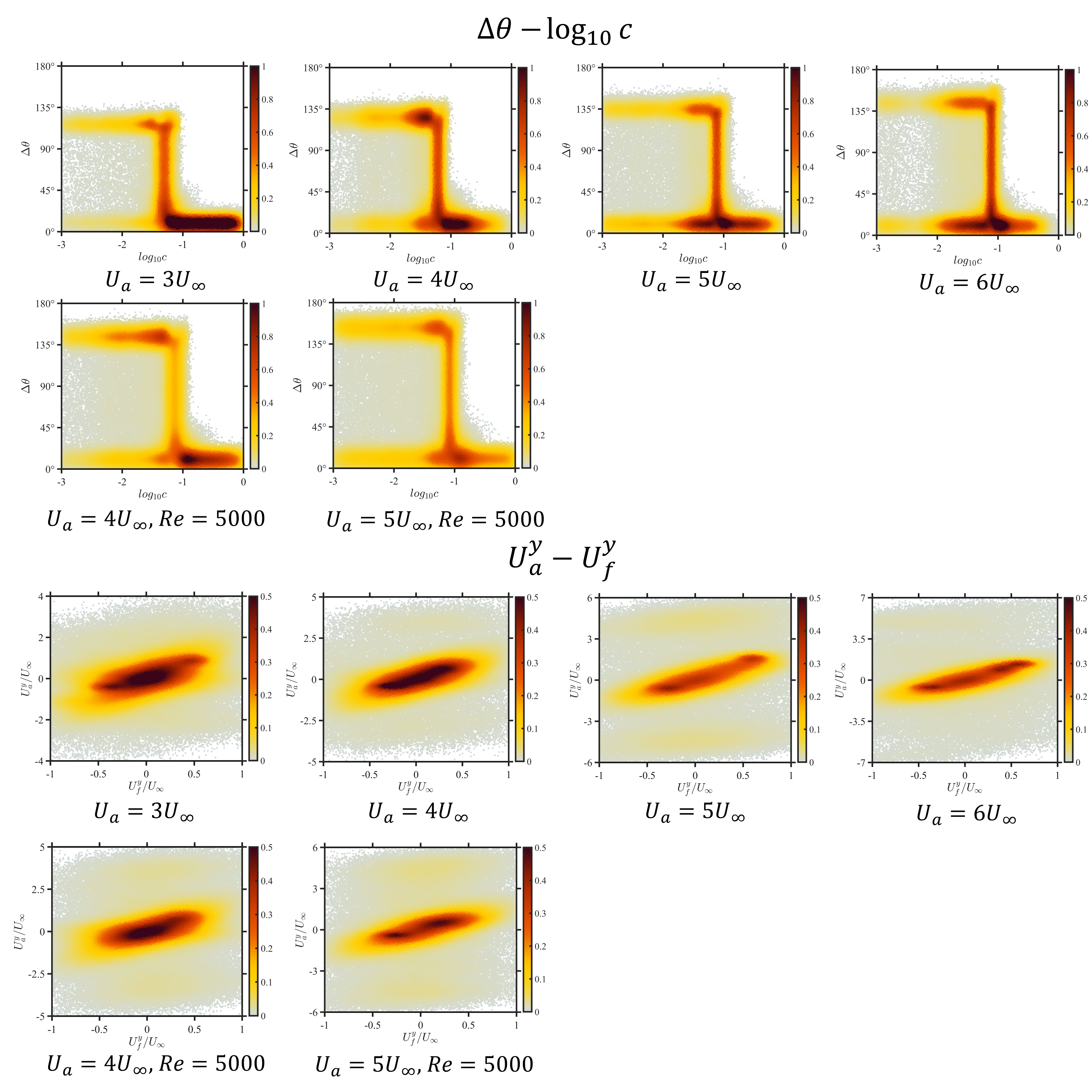}
\caption{Joint probability density distribution of the agent’s state in $\Delta \theta-\log_{10} c$ and $U_{a}^{y}-U_{f}^{y}$ space with optimal memory length $T_M^*$ for different $U_a/U_{\infty}$. The default parameters are $Re=500,Sc=0.1,U_a=3U_{\infty},f=10U_{\infty}/D$.}
\label{fig:supplement_s8}
\end{figure}

\begin{figure}[htbp]
\centering
\includegraphics[width=0.95\linewidth]{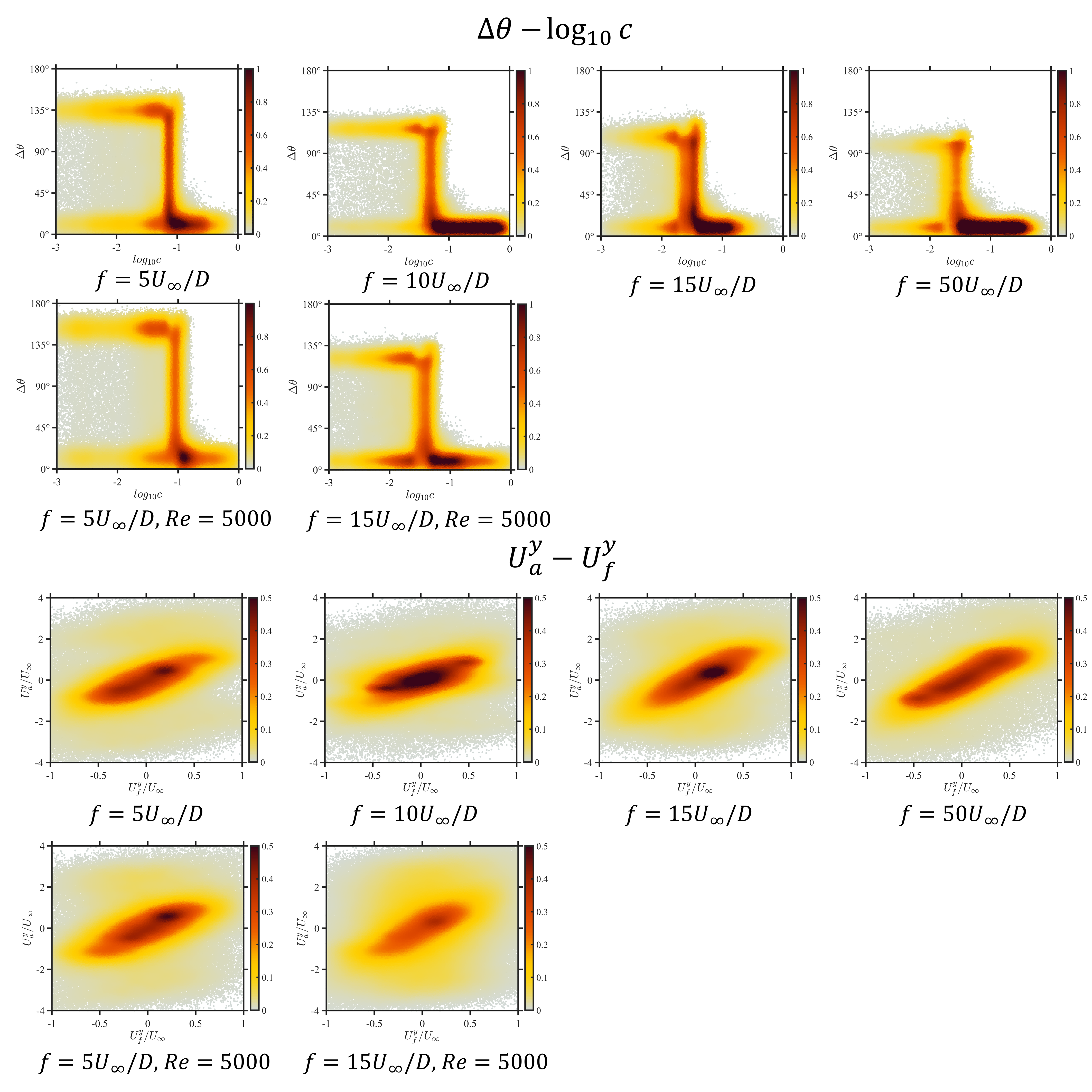}
\caption{Joint probability density distribution of the agent’s state in $\Delta \theta-\log_{10} c$ and $U_{a}^{y}-U_{f}^{y}$ space with optimal memory length $T_M^*$ for different $f$. The default parameters are $Re=500,Sc=0.1,U_a=3U_{\infty},f=10U_{\infty}/D$.}
\label{fig:supplement_s9}
\end{figure}

\quad We further analyze the agent's search strategy under different conditions by plotting its state in $\Delta \theta-\log_{10} c$ and $U_{a}^{y}-U_{f}^{y}$ space. The flow-assisted casting strategy in each case is characterized by the small adjusting angle $\Delta\theta_1$, large casting angle $\Delta\theta_2$, concentration threshold $c_c$, and velocity ratio $k$ obtained from the joint PDF. Figures \ref{fig:supplement_s7} and \ref{fig:supplement_s8} show that these parameters increase monotonically with $Re$, $Sc$, and $U_a/U_{\infty}$. This indicates that the agent learns to cast in a large scanning area, and better exploit the background flow. Conversely, as shown in Figure \ref{fig:supplement_s9}, increasing the detection frequency $f$ decreases $\Delta\theta_1$, $\Delta\theta_2$, and $c_c$, while $k$ increases. In this case, the agent adopts a more conservative strategy, moving straight with greater confidence and executing a narrower search if necessary, which yields a higher search speed.

\subsection{Probability Distribution of $U_{eff}/U_a$ under Different Conditions}

\begin{figure}[htbp]
\centering
\includegraphics[width=0.95\linewidth]{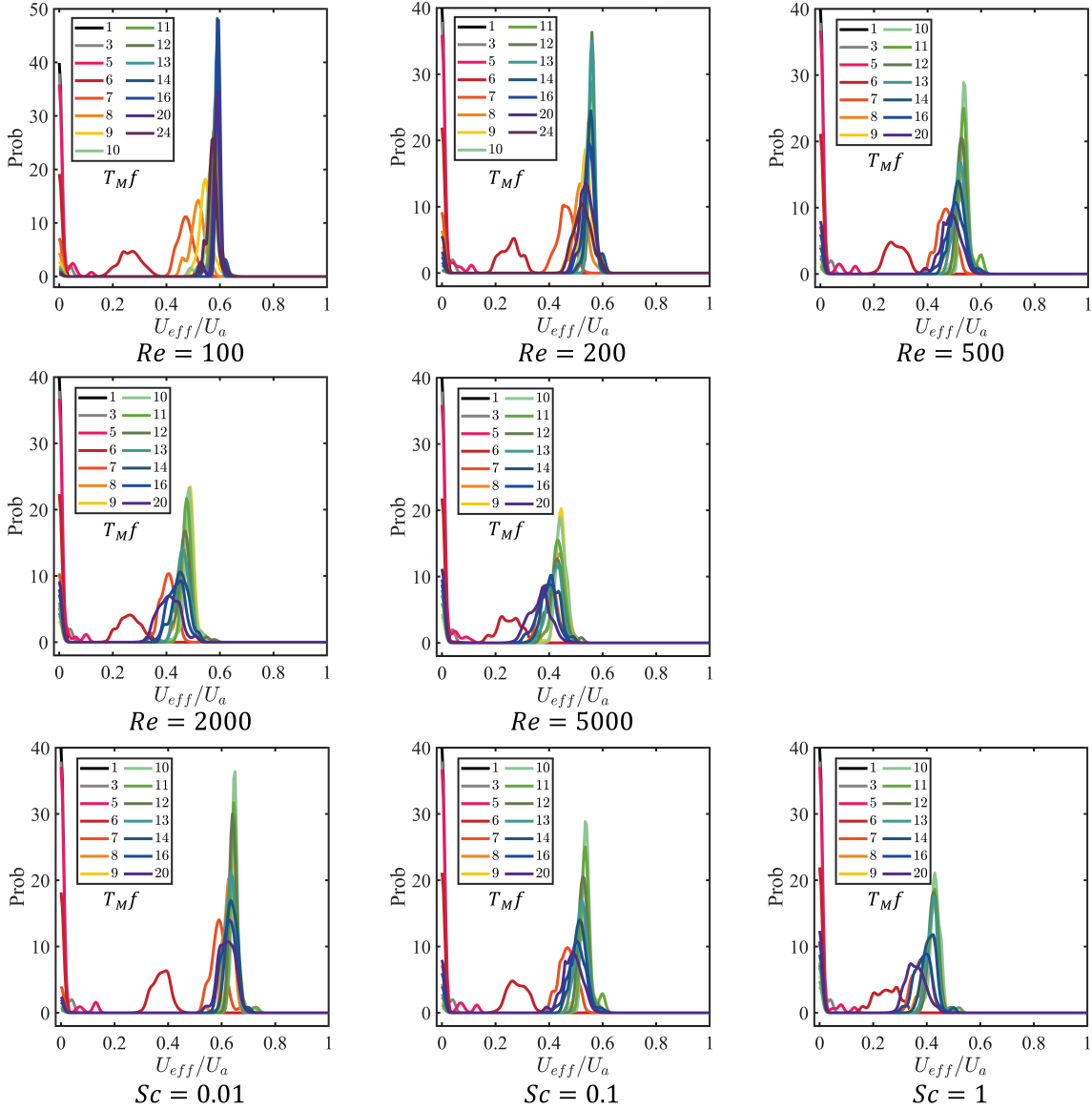}
\caption{The probability distribution of search speed $U_{eff}/U_a$ under different flow conditions $Re$ and $Sc$, with default parameters $Re=500,Sc=0.1,U_a=3U_{\infty},f=10U_{\infty}/D$.}
\label{fig:supplement_s10}
\end{figure}

\begin{figure}[htbp]
\centering
\includegraphics[width=0.95\linewidth]{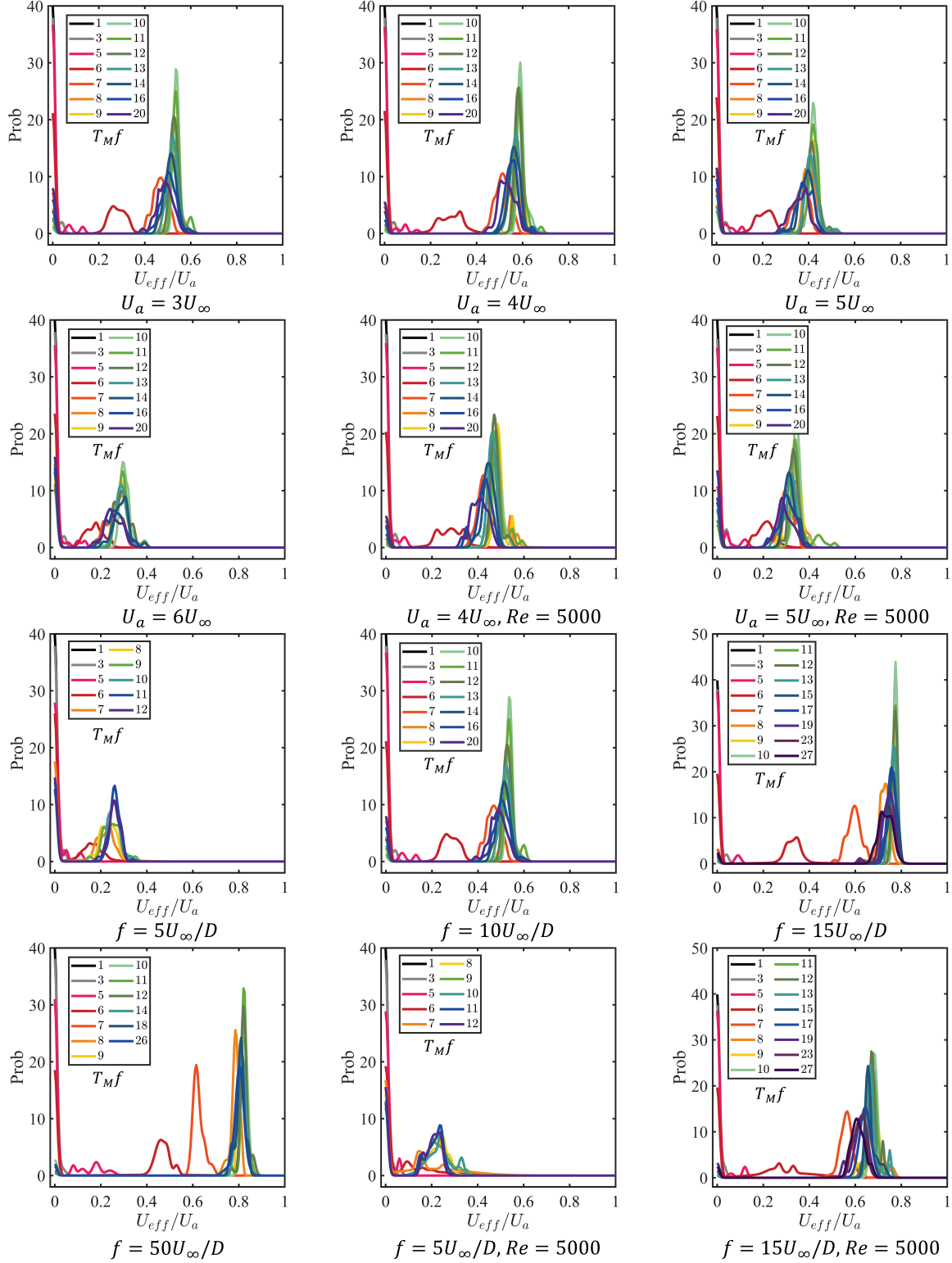}
\caption{The probability distribution of search speed $U_{eff}/U_a$ under different agent configurations $U_a/U_{\infty}$ and $f$, with default parameters $Re=500,Sc=0.1,U_a=3U_{\infty},f=10U_{\infty}/D$.}
\label{fig:supplement_s11}
\end{figure}

\quad The non-monotonic dependence of the agent’s search speed $U_{eff}$ on memory length $T_M$ holds robustly across conditions, as shown by the probability distribution of $U_{eff}/U_a$ in Figures \ref{fig:supplement_s10} and \ref{fig:supplement_s11}. For challenging trials (large $Re$, $Sc$, $U_a/U_{\infty}$, and low $f$), the peak representing successful trials is left-shifted and broader, indicating a lower mean speed and higher variance. For simpler cases (small $Re$, $Sc$, $U_a/U_{\infty}$, and high $f$), the distribution is right-shifted and narrower, reflecting better search performance.

\clearpage

\bibliographystyle{ieeetr}
\bibliography{ref}